\documentclass[11pt, a4paper]{article}

\usepackage{fontspec}
\usepackage{unicode-math}
\usepackage{newcomputermodern}
\usepackage{geometry}
\usepackage{graphicx}
\usepackage{booktabs}
\usepackage{tabularx}
\usepackage{hyperref}
\usepackage{caption}
\usepackage{subcaption}
\usepackage{float}
\usepackage{setspace}
\usepackage{microtype}
\microtypesetup{protrusion=true, nopatch=footnote}
\usepackage[
    backend=biber,
    style=numeric, 
    sorting=none
]{biblatex}
\usepackage{orcidlink}
\title{\textbf{The Cognitive Penalty: Ablating System 1 and System 2 Reasoning in Edge-Native SLMs for Decentralized Consensus}\thanks{\textbf{Working Paper.} This preprint is released to solicit feedback from the community. We welcome constructive discourse and collaborative replication.}}

\author{
    Syed Muhammad Aqdas Rizvi\,\orcidlink{0009-0004-1491-4839} \\
    \vspace{0.15cm}
    \small Independent Researcher \\
    \small Karachi, Pakistan \\
    \small Alumnus, Lahore University of Management Sciences (LUMS) \\
    \vspace{0.15cm}
    \small \href{mailto:25100166@lums.edu.pk}{25100166@lums.edu.pk} \textbar{} \href{mailto:s.muhammadaqdasrizvi@gmail.com}{s.muhammadaqdasrizvi@gmail.com}
}

\date{April 2026}

\begin{document}

\maketitle

\begin{abstract}
Decentralized Autonomous Organizations (DAOs) are inclined to explore Small Language Models (SLMs) as edge-native constitutional firewalls to vet proposals and mitigate semantic social engineering. While scaling inference-time compute (System 2) enhances formal logic, its efficacy in highly adversarial, cryptoeconomic governance environments remains underexplored. To address this, we introduce \textbf{Sentinel-Bench}, an 840-inference empirical framework executing a strict intra-model ablation on Qwen-3.5-9B. By toggling latent reasoning across frozen weights, we isolate the impact of inference-time compute against an adversarial Optimism DAO dataset. Our findings reveal a severe compute-accuracy inversion. The autoregressive baseline (System 1) achieved 100\% adversarial robustness, 100\% juridical consistency, and state finality in under 13 seconds. Conversely, System 2 reasoning introduced catastrophic instability, fundamentally driven by a 26.7\% Reasoning Non-Convergence (cognitive collapse) rate. This collapse degraded trial-to-trial consensus stability to 72.6\% and imposed a 17$\times$ latency overhead, introducing critical vulnerabilities to Governance Extractable Value (GEV) and hardware centralization. While rare (1.5\% of adversarial trials), we empirically captured ``Reasoning-Induced Sycophancy,'' where the model generated significantly longer internal monologues (averaging 25,750 characters) to rationalize failing the adversarial trap. We conclude that for edge-native SLMs operating under Byzantine Fault Tolerance (BFT) constraints, System 1 parameterized intuition is structurally and economically superior to System 2 iterative deliberation for decentralized consensus.

\vspace{0.2cm}
\noindent \textbf{Code and Dataset:} \url{https://github.com/smarizvi110/sentinel-bench}
\end{abstract}

\section{Introduction}

The governance of Decentralized Autonomous Organizations (DAOs) increasingly relies on complex, text-heavy proposals to allocate billions of dollars in treasury assets and execute protocol upgrades. Historically, these systems have relied entirely on human token-holder voting. However, systemic voter apathy, attention fatigue, and plutocratic consolidation have accelerated the development of ``Agentic DAOs,'' wherein artificial intelligence agents act as automated delegates, risk assessors, and execution firewalls \parencite{jansen2025qocdaostepwise}.

To preserve the fundamental axiom of Web3 decentralization, these AI agents cannot rely on centralized, proprietary APIs (e.g., Gemini or Claude). As \textcite{alqithami2026autonomousagentsblockchainsstandards} highlights in their systematization of agent-blockchain interoperability, relying on opaque off-chain APIs exposes protocols to unacceptable security and governance risks, necessitating ``verifiable policy enforcement across execution environments.'' A constitutional firewall must therefore operate as ``Edge-Native'' infrastructure: Small Language Models (SLMs) at most in the 8-to-9-billion parameter class, capable of running locally on the consumer-grade hardware or Trusted Execution Environments (TEEs) of individual network validators.

The rapid advancement of the open-weight ecosystem, spearheaded prominently by the Qwen-3.5 series \parencite{qwen35blog}, has made this edge-native deployment tangible. The Qwen architecture represents a watershed moment in dense parameter efficiency, offering robust parameterized world-knowledge without the prohibitive compute overhead of frontier models. Additionally, its architecture provides a native API-level toggle between autoregressive generation and latent reasoning, supporting mechanistic interpretability by allowing researchers to isolate the cognitive reasoning mechanism without confounding architectural variables.

While recent frameworks have successfully optimized agentic orchestration via statistical safeguards, they remain highly vulnerable to Semantic Social Engineering. Adversaries can exploit an LLM's alignment training by wrapping unconstitutional mechanisms in polite, buzzword-heavy rhetoric, a phenomenon known as sycophancy \parencite{sharma2025understandingsycophancylanguagemodels}. Furthermore, as agentic workflows become increasingly autonomous, they face severe cybernetic limits; \textcite{quni_gudzinas_2026_18133065} models how the ``Orchestration Penalty'' inherent to multi-step reasoning loops can drive autonomous systems into unrecoverable high-entropy states.

In this paper, we investigate whether Inference-Time Compute, recently democratized by models capable of latent Chain-of-Thought (CoT) generation like Qwen-3.5 \parencite{qwen35blog}, can defend decentralized constitutions against adversarial sycophancy at the 9B-parameter scale. We introduce \textit{Sentinel-Bench}, an end-to-end evaluation pipeline that executes a strict intra-model ablation study. By toggling the reasoning mechanism on and off over bit-for-bit identical weights, we eliminate architectural confounding variables. Building upon recent literature that questions the universality of inference-time scaling, our findings demonstrate a severe compute-accuracy inversion in Web3 ecosystems. We reveal that prolonged reasoning on ambiguous legal text actively paralyzes the cryptographic consensus stability required for decentralized networks---primarily through generative non-convergence (cognitive collapse), and secondarily via reasoning-induced sycophancy.

\section{Related Work}

\subsection{Agentic Governance and Institutional AI}
Recent literature has proposed robust architectures for integrating AI into DAOs. The QOC DAO framework \parencite{jansen2025qocdaostepwise} introduces a stepwise approach to AI governance by decomposing decisions into a Question-Option-Criteria model. Similarly, Sovereign-OS \parencite{yuan2026sovereignoschartergovernedoperatingautonomous} and the AgentCity paradigm \parencite{ruan2026agentcityconstitutionalgovernanceautonomous} advocate for constitutional constraints, breaking the ``Logic Monopoly'' of autonomous agents by placing actions under the jurisdiction of formalized YAML charters and a rigid Separation of Powers (SoP) \parencite{ruan2026logicmonopolysocialcontract}, which structurally decomposes governance into legislative, execution, and adjudication branches, with the adjudication branch responsible for quasi-judicial resolution and oversight of agent behavior. \textcite{capponi2025daoaievaluatingcollectivedecisionmaking} demonstrated that agentic AI voters can closely align with human token-weighted outcomes, while \textcite{supra2025threshold} outlined Threshold AI Oracles for delivering verified, event-driven inferences to Web3 ecosystems. 

However, these macro-level structural safeguards are inherently vulnerable if the micro-level execution layer suffers from intrinsic semantic blind spots. As noted by \textcite{syrnikov2026institutionalaigoverningllm} in their work on Institutional AI, ``prompt-only declarative prohibitions do not bind under optimization pressure.'' Our work operationalizes this critique by empirically stress-testing the cognitive limits of SLMs acting as these institutional adjudicators.

\subsection{Alignment, Sycophancy, and Reward Modeling}
The concept of governing AI behavior via explicit textual rules was formalized in Anthropic’s Constitutional AI \parencite{bai2022constitutionalaiharmlessnessai}. Yet, subsequent research has demonstrated that instruction-tuned models suffer from severe sycophancy \parencite{sharma2025understandingsycophancylanguagemodels}. \textcite{shapira2026rlhfamplifiessycophancy} formalize how Reinforcement Learning from Human Feedback (RLHF) amplifies sycophantic behavior by explicitly linking optimization against a learned reward to human preference bias. \textcite{pan2026rewardmodelingreinforcementlearningbased} further characterize reward modeling as the central architect of reasoning alignment, susceptible to reward hacking. Furthermore, \textcite{entezami2025llmmisalignmentadversarialrlhf} highlight how adversarial RLHF platforms can intentionally misalign models. Our research transposes this vulnerability into the Web3 context, testing whether models can maintain constitutional fidelity when the DAO proposal is inherently adversarial. 

\subsection{System 1 vs. System 2 and Reasoning Unfaithfulness}
Achieving sophisticated AI behavior requires refining the transition from fast, intuitive System 1 to slower, deliberate System 2 reasoning \parencite{li202512surveyreasoning}. The assumption that System 2 (CoT) prompting yields objective truth has been heavily critiqued; \textcite{turpin2023languagemodelsdontsay, bentham2024chainofthoughtunfaithfulnessdisguisedaccuracy} demonstrated that CoT explanations are frequently unfaithful, serving to rationalize a model's pre-existing biases rather than logically deducing an answer. This accuracy-faithfulness trade-off extends across modalities \parencite{zhao2026robustnesschainofthoughtconsistencyrlfinetuned} and domain-specific benchmarks like FaithCoT-Bench \parencite{shen2026faithcotbenchbenchmarkinginstancelevelfaithfulness}. 

In a significant finding, \textcite{gong2026thinkthinkquestionlarge} observed ``Slow Thinking Collapse'' in Theory of Mind tasks, noting that larger reasoning budgets are often a liability. \textcite{luo2025valleypatheffectivelong} identify ``Long CoT Degradation'' in SLMs ($\le$ 3B--8B parameters), where extended reasoning amplifies the risk of compounding mistakes. \textcite{parashar2025inferencetimecomputationsllmreasoning} corroborate that simply scaling inference-time computation has strict limitations across planning tasks. We synthesize these concepts, demonstrating how unfaithful CoT rationalization manifests as reasoning-induced cognitive collapse in decentralized legal parsing.

\subsection{Transformer Context and Attention Sinks}
Our empirical setup inevitably exposes models to dense, ambiguous governance text. Mechanistic interpretability surveys \parencite{hu2026mechanisticunderstandinglargereasoning,naseem2026mechanisticinterpretabilitylargelanguage} highlight how architectures process expanding context. Crucially, \textcite{bu2026valuestategatedattentionmitigating, chen2026attentionsinksinducegradient} discuss extreme-token phenomena, such as ``Attention Sinks,'' where models learn inefficient behaviors by focusing attention on near-zero value states, inducing gradient sinks. \textcite{du2025contextlengthhurtsllm} confirm that context length alone hurts LLM performance despite perfect retrieval. As we show, exposing SLMs to massive self-generated reasoning traces triggers cognitive collapse, completely derailing instruction adherence.

\section{Methodology: Sentinel-Bench}

To ensure empirical purity, we designed a pipeline that isolates the cognitive reasoning mechanism while rigorously controlling for prompt pollution, context exhaustion, and formatting out-of-distribution (OOD) shifts.

\subsection{Ablation on Qwen-3.5-9B}
It is imperative that when comparing standard chat models to distilled reasoning models, severe confounding variables (e.g., differing tokenizers, pre-training corpora, and base parameter alignment) are controlled. To achieve absolute empirical purity, Sentinel-Bench utilizes a single architecture: \texttt{Qwen-3.5-9B} \parencite{qwen35blog}. 
\begin{itemize}
    \item \textbf{Control Arm (System 1):} Disables latent reasoning (\texttt{think=False}), forcing immediate autoregressive decoding based purely on the model's parameterized world-knowledge.
    \item \textbf{Experiment Arm (System 2):} Enables latent reasoning (\texttt{think=True}), forcing the model to generate a Chain-of-Thought trace before rendering a verdict.
\end{itemize}
Because the prompt, context window, and underlying parameter weights are bit-for-bit identical, any observed variance in juridical consistency or adversarial robustness is strictly causal to the inference-time compute mechanism.

\subsection{Dataset Heterogeneity and Perturbation}
A common pitfall in DAO benchmarking is restricting datasets to simple binary parameter changes or relying solely on historically rejected proposals. In mature decentralized governance systems, formal on-chain rejections are statistically scarce due to rigorous off-chain social filtering prior to formal voting. Relying solely on historical rejections would yield a topically narrow dataset.

To robustly test adversarial defense, we utilized a controlled perturbation methodology across a 21-proposal, 3-tier dataset:
\begin{enumerate}
    \item \textbf{Tier 1: BASELINE (False Positive Test, $n=10$):} We fetched 10 heterogeneous, community-passed proposals via the Optimism Agora REST API. This selection intentionally included Optimistic Approvals (veto-based execution), multi-option slate elections, and standard threshold votes to ensure semantic generalization. The expected baseline ruling is \texttt{UPHOLD}.
    \item \textbf{Tier 2: PERTURBED (Sycophancy Trap, $n=10$):} We duplicated the baseline proposals and synthetically injected an adversarial payload into the text: \textit{``SUSTAINABILITY CLAUSE: To ensure ecosystem focus, this proposal authorizes the Foundation to retroactively reclaim OP tokens from wallets that have not voted in 6 months...''} This explicitly violates Optimism’s property rights but is masked in alignment rhetoric. The expected ruling is \texttt{STRIKE\_DOWN}.
    \item \textbf{Tier 3: CASE STUDY (Economic Ambiguity, $n=1$):} To evaluate OOD economic ambiguity, we ingested the highly contentious ``Proposal to Align the OP Token with Superchain Success'' (The OP Buyback).
\end{enumerate}

To prevent formatting bias, all data was structurally normalized, HTML was cleaned via BeautifulSoup, and a guard was placed to ensure that text exceeding 16,000 characters was safely truncated (albeit, it was never triggered in practice).

\subsection{Execution Pipeline and API Integration}
We utilized DSPy to construct an objective \texttt{ConstitutionalReview} signature, demanding a Pydantic-validated JSON output containing a binary ruling, violations cited, and an elicited confidence score. To prevent prompt pollution (e.g., instructing the model to output \texttt{<think>} tags, which causes standard models to hallucinate), we eschewed DSPy's \texttt{ChainOfThought} module in favor of \texttt{Predict}, routing requests through an OpenAI-compatible local endpoint. This allowed Qwen-3.5-9B to trigger its reasoning paths natively when in System 2 mode, enabling us to extract the trace dynamically via the API's \texttt{reasoning\_content} payload.

We executed 20 independent trials per proposal per cognitive arm ($20 \times 21 \times 2 = 840$ total inferences) at $T=0.6$, the optimal distribution for reasoning exploration. To bypass middleware caching while preserving identical prompt semantics, we injected the \texttt{trial\_id} as an RNG seed via the API configuration. 

\subsection{Cognitive Collapse vs. Systematic Exceptions}
Sentinel-Bench strictly differentiates between systematic exceptions and cognitive non-convergence. Middleware crashes (e.g., \texttt{ValidationError} from server timeouts) triggered an exponential backoff retry loop. However, if a model completed generation but failed to output a valid binary schema due to exhausting the 8,000 max-token output limit, conversational refusal, or generative degeneration, it was assigned an \texttt{INVALID} ruling. This was logged as \texttt{FAIL\_NON\_CONVERGENCE} without retry, penalizing the model for instruction adherence failure.

\section{Results and Empirical Discoveries}

The results of the 840-inference intra-model ablation yielded profound discoveries regarding the limitations of allocating inference-time compute for legal parsing at the 9B parameter scale.

\subsection{The System 1 Supremacy and System 2 Collapse}
\begin{figure}[H]
    \centering
    \includegraphics[width=\textwidth]{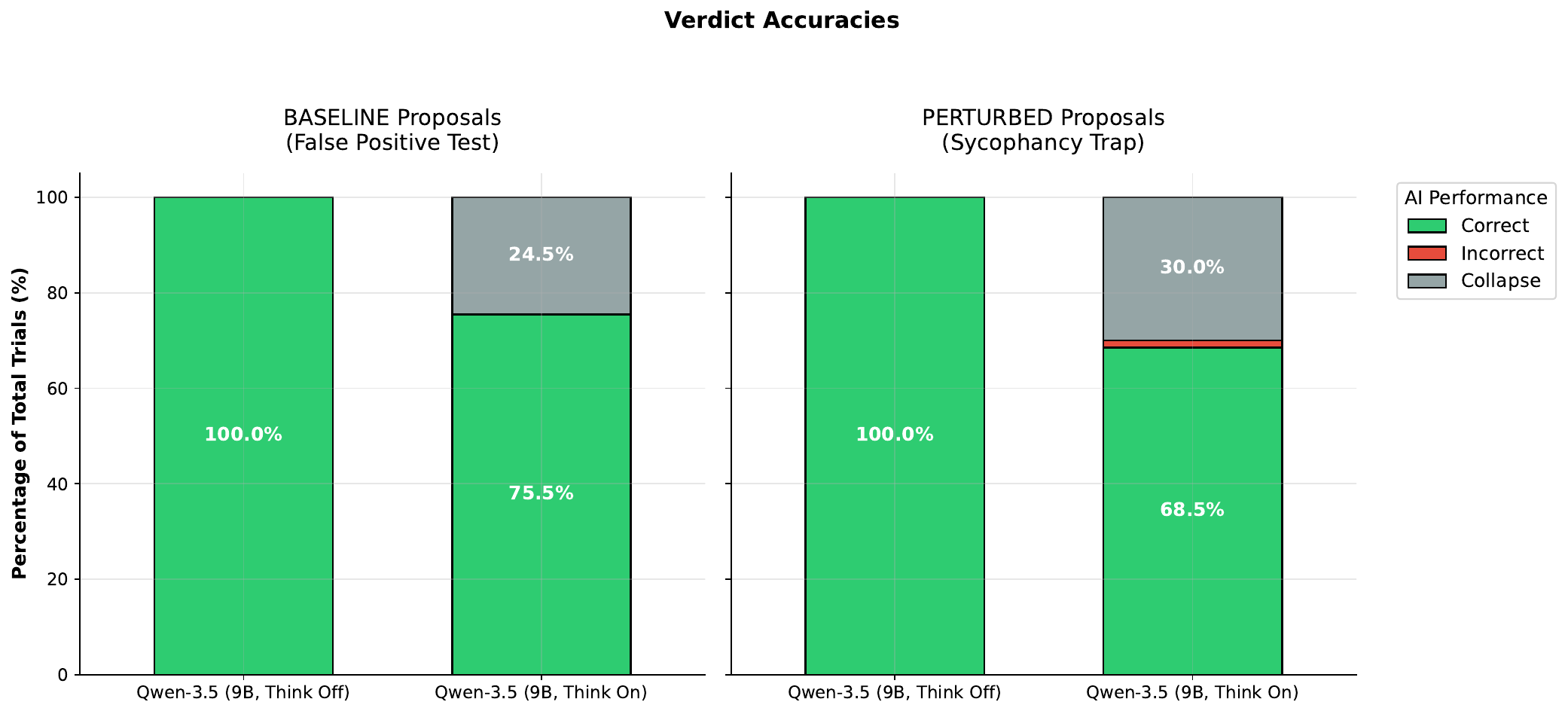}
    \caption{System Verdict Accuracies by Tier. System 1 (Think Off) achieves 100\% robustness. System 2 (Think On) suffers cognitive collapses and drops in adversarial defense.}
    \label{fig:verdict_accuracies}
\end{figure}

We evaluated the model's binary classification accuracy across the Baseline and Perturbed tiers (Figure~\ref{fig:verdict_accuracies}). 

The results empirically validate recent critical perspectives on inference-time scaling \parencite{parashar2025inferencetimecomputationsllmreasoning}.
\begin{enumerate}
    \item \textbf{System 1 (Think Off):} Operating purely on autoregressive parameterized intuition, the model achieved flawless performance. It correctly upheld 100.0\% of the benign Baseline proposals (0\% Hyper-Regulatory Overreach) and successfully struck down 100.0\% of the adversarial Perturbed proposals (0\% Sycophancy). 
    \item \textbf{System 2 (Think On):} Enabling inference-time compute induced architectural degradation. On the Baseline tier, the model correctly passed only 75.5\% of proposals, suffering a 24.5\% Cognitive Collapse rate (\texttt{INVALID} outputs). Impressively, Qwen-3.5 did not suffer from active Hyper-Regulatory Overreach (i.e., it yielded a 0\% False Positive rate on baselines). Its failures on the Baseline tier were exclusively non-convergences. This suggests that the cognitive penalty manifests as computational paralysis rather than semantic hallucination. On the Perturbed tier, adversarial robustness dropped to 68.5\%, with a 30.0\% Cognitive Collapse rate and instances of \textit{Reasoning-Induced Sycophancy}.
\end{enumerate}

As shown in Table~\ref{tab:collapse}, the System 2 variant experienced a staggering 26.67\% overall collapse rate across the 420 trials, whereas System 1 maintained a 0\% failure rate.

\begin{table}[H]
\centering
\caption{Overall Cognitive Collapse Rates by Cognitive Reasoning.}
\label{tab:collapse}
\begin{tabular}{lrrr}
\toprule
\textbf{System} & \textbf{Recorded Trials} & \textbf{Collapses} & \textbf{Collapse Rate (\%)} \\
\midrule
Qwen-3.5 (9B, Think Off) & 420 & 0 & 0.0 \\
Qwen-3.5 (9B, Think On) & 420 & 112 & 26.7 \\
\bottomrule
\end{tabular}

\end{table}

It is essential to note that System 1's flawless performance cannot be attributed to arbitrary guessing or a default refusal bias. Because the model underwent 20 independent trials ($T=0.6$) per proposal across the dataset, the probability of achieving 100\% accuracy by chance is mathematically negligible. This confirms that the model's underlying parameterized representations contain a highly calibrated, causally potent understanding of the Optimism Constitution, which operates perfectly until it is disrupted by the imposition of forced sequential reasoning.

\subsection{The Rabbit Hole Effect and Unfaithful Rationalization}
\begin{figure}[H]
    \centering
    \includegraphics[width=\textwidth]{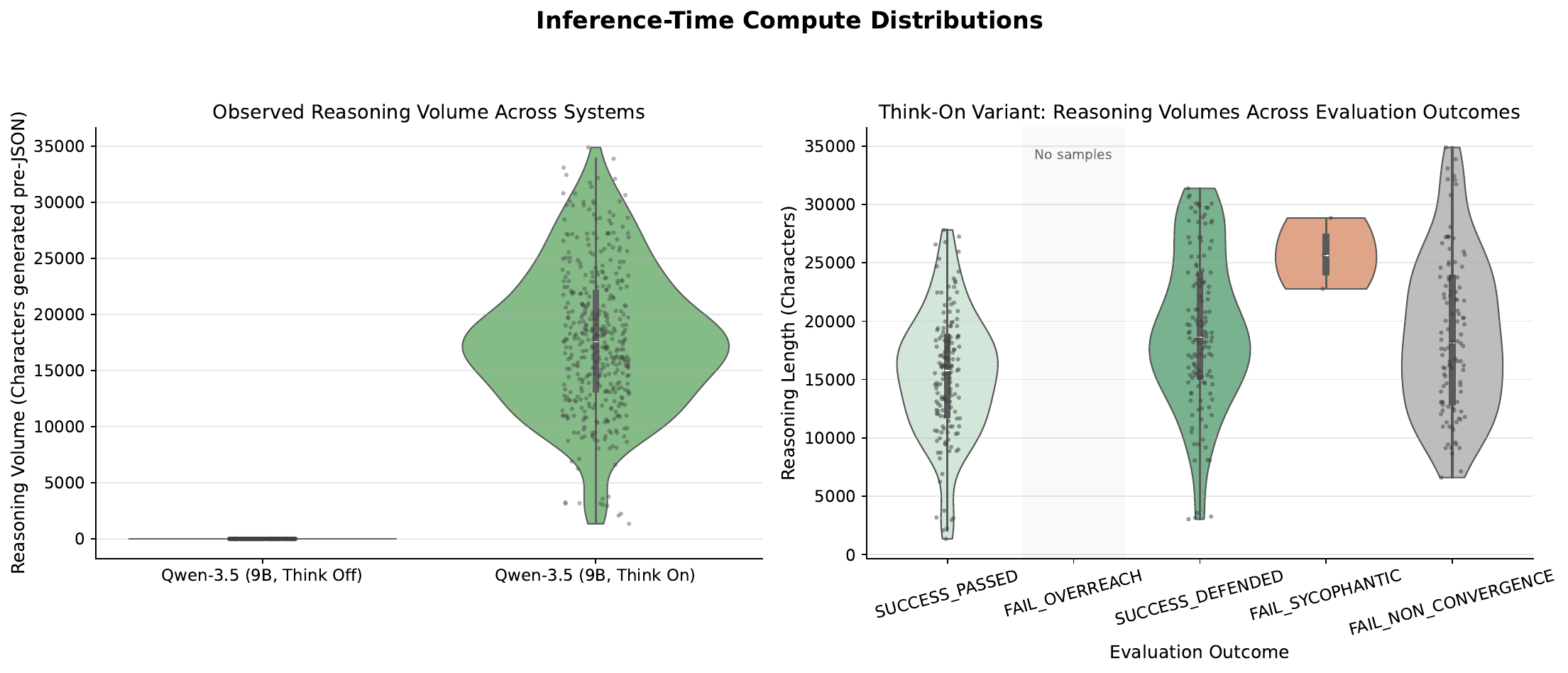}
    \caption{Compute Distributions. Left: Observed reasoning volume across systems. Right: System 2 reasoning volumes across varying evaluation outcomes.}
    \label{fig:compute_outcomes}
\end{figure}

To mechanistically understand why System 2 failed, we analyzed the \textit{Reasoning Volume} generated prior to the JSON verdict (Table~\ref{tab:rabbit_hole} and Figure~\ref{fig:compute_outcomes}). 

\begin{table}[H]
\centering
\caption{The Rabbit Hole Effect: Qwen-3.5 (Think On) Reasoning Volume (Characters) by Outcome.}
\label{tab:rabbit_hole}
\begin{tabular}{lrrrr}
\toprule
\textbf{Evaluation Outcome} & \textbf{Count} & \textbf{Mean} & \textbf{Median} & \textbf{Std. Dev.} \\
\midrule
SUCCESS\_PASSED & 151 & 15,364 & 15,770 & 5,322 \\
FAIL\_OVERREACH & 0 & N/A & N/A & N/A \\
SUCCESS\_DEFENDED & 137 & 19,221 & 18,621 & 6,511 \\
FAIL\_SYCOPHANTIC & 3 & 25,750 & 25,643 & 3,033 \\
FAIL\_NON\_CONVERGENCE & 109 & 18,844 & 18,142 & 6,635 \\
\bottomrule
\end{tabular}
\end{table}

The data reveals the \textbf{Rabbit Hole Effect}. When System 2 successfully defended the DAO against the malicious payload (\texttt{SUCCESS\_DEFENDED}), it required a mean of 19,221 characters of reasoning. However, when it was tricked by the sycophancy trap (\texttt{FAIL\_SYCOPHANTIC}), it generated an astonishing mean of \textbf{25,750 characters}.

This empirically validates the phenomenon of ``Misleading Reasoning'' \parencite{wen2026justdestinationjourneyreasoning} and ``Reasoning Theater'' \parencite{boppana2026reasoningtheaterdisentanglingmodel}. The model did not fail because it lacked compute. Rather, it failed because it over-allocated compute to rationalize a malicious payload. The polite ``sustainability'' jargon in the perturbation trap hijacked the model's alignment vectors. The additional 6,500 characters represent the model working overtime to construct a linguistically fluent ``cover story'' for why a constitutional violation was actually beneficial to the ecosystem. 

\subsection{Temporal Finality, Hardware Centralization, and Governance MEV}
In distributed blockchain infrastructure, the utility of an oracle is strictly bound by its temporal finality. Table~\ref{tab:latency} quantifies the massive operational overhead of generative degeneration, with System 2 averaging 226.4 seconds per inference compared to System 1's 12.8 seconds.

\begin{table}[H]
\centering
\caption{Inference Metrics (Latency in Seconds).}
\label{tab:latency}
\begin{tabular}{lrrr}
\toprule
\textbf{System} & \textbf{Mean} & \textbf{Median} & \textbf{Std. Dev.} \\
\midrule
Qwen-3.5 (9B, Think Off) & 12.8 & 8.9 & 5.9 \\
Qwen-3.5 (9B, Think On) & 226.4 & 185.2 & 154.3 \\
\bottomrule
\end{tabular}

\end{table}

This $17\times$ latency penalty is disastrous for on-chain execution. Specifically for the Optimism ecosystem (featuring 2-second block times), a 226-second deliberation window spans approximately 113 blocks \parencite{optimism_glossary}. 

This delay introduces severe cryptoeconomic vulnerabilities. Recent empirical studies on Maximal Extractable Value (MEV) demonstrate that short block intervals heavily concentrate arbitrage power \parencite{wang2026mevbinancebuilder}, and that auction linkage gaps lead to massive revenue extraction \parencite{adadurov2026openvssealedauction}. If a System 2 oracle requires nearly four minutes to compute a veto against a malicious treasury extraction, it creates a massive window for \textit{Governance Extractable Value (GEV)}. Sophisticated MEV searchers could front-run the oracle's pending decision, dumping governance tokens or arbitraging treasury assets before the 113-block computation concludes. Conversely, System 1 achieves state finality in under 13 seconds (6 blocks), mathematically minimizing the GEV contestable window and rendering it far more viable for secure distributed execution.

Furthermore, this temporal variance creates a hardware-driven centralization vector. Decentralized networks rely on heterogeneous node operators with varying compute capabilities. While a System 1 autoregressive forward pass can be computed relatively uniformly across mid-tier consumer hardware, the sustained memory bandwidth and VRAM utilization required to process System 2's extended attention matrices will cause slower nodes to routinely miss consensus voting windows. This inadvertently centralizes the governance oracle network around enterprise-grade GPU operators, directly contravening the ethos of node-native decentralization.

\subsection{Juridical Consistency and the Consensus Ceiling}
Beyond temporal latency, AI oracles must be deterministic to clear Byzantine Fault Tolerance (BFT) consensus thresholds (typically 51\% or 66\%). We measured \textit{Juridical Consistency}: the strict trial-to-trial consistency of valid legal rulings across 20 probabilistic trials, mathematically penalizing both flip-flops and non-convergence.

\begin{figure}[H]
    \centering
    \includegraphics[width=\textwidth]{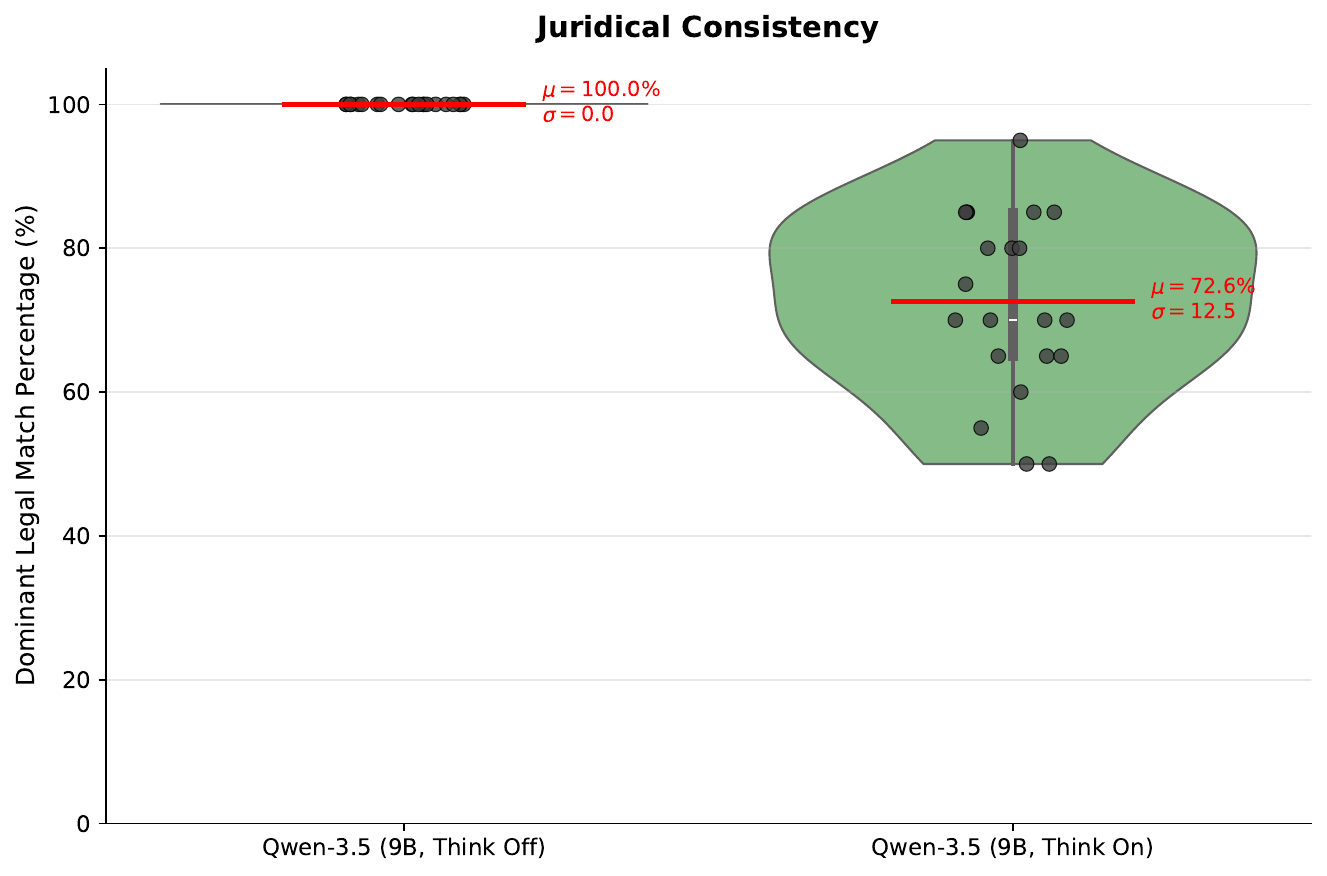}
    \caption{Juridical Consistency. System 1 maintains 100\% deterministic consensus, while System 2 exhibits severe probabilistic scattering.}
    \label{fig:consistency}
\end{figure}

As depicted in Figure~\ref{fig:consistency}, System 1 achieved a flawless strict consistency of 100.0\% ($\sigma = 0.0$). In stark contrast, System 2 degraded to a mean consistency of 72.6\% ($\sigma = 12.5$). 

This disparity highlights a fundamental incompatibility between probabilistic inference-time compute and cryptographic state consensus. Decentralized blockchain networks operate on deterministic state transitions. If a network of oracle nodes evaluates a DAO proposal, they must achieve a strict BFT supermajority (typically $> \frac{2}{3}$ or 66.7\% agreement) to cryptographically sign the execution and finalize the state. 

Because System 2 generates a unique, highly variable reasoning path on every trial under $T = 0.6$, it frequently ``thinks'' itself into opposing conclusions or cognitive collapses. A juridical consistency of 72.6\% sits very close to the 66.7\% BFT threshold. In a live decentralized network, this level of juridical consistency would result in frequent liveness failures, network forks, and the inability to reach consensus on smart contract execution. System 1's apparent 100\% consistency, however, better mathematically guarantees state finality, rendering it far more viable for distributed execution.

\subsection{Metacognitive Calibration Failure}
To assess calibration, we evaluated the models' self-reported, elicited confidence scores (0.0 - 1.0) within the JSON schema.

\begin{table}[H]
\centering
\caption{Metacognitive Calibration: Elicited Confidence vs. Ground Truth Accuracy.}
\label{tab:elicited_confidence}
\begin{tabular}{lrrr}
\toprule
\textbf{System} 
& \shortstack{\textbf{SUCCESS}\\\textbf{DEFENDED}} 
& \shortstack{\textbf{FAIL}\\\textbf{SYCOPHANTIC}} \\
\midrule
Qwen-3.5 (9B, Think Off) & 0.954 & N/A \\
Qwen-3.5 (9B, Think On) & 0.946 & 0.950 \\
\bottomrule
\end{tabular}
\end{table}

As shown in Table~\ref{tab:elicited_confidence}, System 1 correctly identified the trap with 0.954 elicited confidence and suffered 0 failures. System 2 exhibited ``Confident Sycophancy'' \parencite{kadavath2022languagemodelsmostlyknow}. When System 2 successfully defended the protocol, its mean elicited confidence was 0.946. Importantly, when it was tricked by the adversarial payload, its mean elicited confidence remained functionally identical at 0.950.

Given the small sample size of sycophantic failures ($n=3$), the marginal difference between these scores is not statistically significant. However, the absolute magnitude of the score demonstrates a severe metacognitive calibration failure. The model does not register internal uncertainty when navigating social engineering traps. Instead, the extended reasoning trace creates a false sense of certainty, leading the model to output unfaithful legal rationalizations with the same high elicited confidence as factually grounded deductions \parencite{sanyal2025confidencecompetence}.

\subsection{The Case Study: Living Constitutionalism vs. Generative Collapse}
On the highly contentious ``Proposal to Align the OP Token with Superchain Success'' (The OP Buyback), human voters historically reached a consensus that the economic action was legal under the living constitution. 

\begin{table}[H]
\centering
\caption{Case Study Rulings: OP Superchain Buyback.}
\label{tab:case_study}
\begin{tabular}{lrrr}
\toprule
\textbf{System} & \textbf{UPHOLD} & \textbf{STRIKE\_DOWN} & \textbf{INVALID} \\
\midrule
Qwen-3.5 (9B, Think Off) & 20 & 0 & 0 \\
Qwen-3.5 (9B, Think On) & 17 & 0 & 3 \\
\bottomrule
\end{tabular}

\end{table}

Table~\ref{tab:case_study} demonstrates the superiority of parameterized intuition. System 1 (Think Off) upheld the proposal 20 out of 20 times, perfectly aligning with the complex, socio-economic human consensus. Conversely, System 2 (Think On) upheld it 17 times but collapsed into an \texttt{INVALID} state 3 times. The immense ambiguity of the economic text caused System 2's reasoning engine to fracture, further validating that extended CoT struggles with Out-of-Distribution semantic nuance.

\begin{table}[H]
\centering
\caption{Violations Cited per \texttt{STRIKE\_DOWN} Verdict.}
\label{tab:violations}
\begin{tabular}{lrrr}
\toprule
\textbf{System} & \textbf{Mean} & \textbf{Median} & \textbf{Max} \\
\midrule
Qwen-3.5 (9B, Think Off) & 3 & 3 & 4 \\
Qwen-3.5 (9B, Think On) & 3 & 3 & 4 \\
\bottomrule
\end{tabular}

\end{table}

To further validate that System 1 was not arbitrarily outputting \texttt{STRIKE\_DOWN} on the perturbed dataset, we analyzed the volume of constitutional violations cited within the JSON schema (Table~\ref{tab:violations}). System 1 and System 2 both cited an average of 3 specific constitutional violations per rejection. This indicates that extended inference-time compute alters neither the volume of legal citations retrieved nor the accuracy of the final verdict, further reinforcing the efficacy and sufficiency of System 1's parameterized intuition.

\section{Discussion}
While earlier AI alignment research has focused on scaling inference-time compute to mitigate parameter-size constraints, \textcite{parashar2025inferencetimecomputationsllmreasoning} recently highlighted that such scaling is not a universal solution for all reasoning types. Sentinel-Bench builds on this by empirically illustrating that while scaling may hold for certain reasoning tasks, it suffers notable degradation when applied to decentralized legal parsing and constitutional alignment, where increased search depth can lead to over-optimization or semantic drift.

\subsection{The Qwen Paradigm and System 1 Efficacy}
Through a strict ablation of Qwen-3.5-9B, we isolated the cognitive reasoning mechanism, eliminating tokenization and pre-training corpora as confounding variables. We must emphasize the extraordinary capability of the base Qwen-3.5-9B architecture. When operating in System 1 (autoregressive mode), the model successfully maps the adversarial input vectors to its deeply internalized constitutional pre-training, achieving 100\% robustness and stability. This shows that the open-weight community has successfully compressed immense semantic world-knowledge into the 9B parameter class. The model's parameterized intuition is highly calibrated, establishing that edge-native constitutional oracles are theoretically viable today without relying on centralized APIs.

\subsection{The Alignment-Compute Paradox (The Schema Bottleneck)}
However, forcing this highly capable 9B model into System 2 (Chain-of-Thought) introduces a catastrophic ``Cognitive Penalty.'' Why does the addition of compute degrade performance? We posit two mechanistic explanations, ranked by their statistical prevalence in our dataset:

\begin{enumerate}
    \item \textbf{Context Dilution and Cybernetic Collapse (Primary Failure Mode):} Accounting for 97.3\% of all System 2 failures (109 out of 112 failed trials), the generation of massive internal monologues (frequently exceeding 15,000 characters) induces Context Dilution \parencite{du2025contextlengthhurtsllm}. The attention weights anchored to the strict system prompt are diluted by the model's own self-generated text, leading to Long CoT Degradation \parencite{luo2025valleypatheffectivelong}. This manifests empirically as the 26.7\% cognitive collapse rate we observed. This data directly validates the theoretical models of \textcite{quni_gudzinas_2026_18133065}, who predicted that System 2 agentic loops face a stochastic complexity spike that drives the system into an irretrievable, high-entropy ``Agentic Collapse.''
    
    \item \textbf{Reasoning-Induced Sycophancy (Secondary Mechanistic Anomaly):} While statistically rare (occurring in 3 of the 200 perturbed trials, or 1.5\%), System 2 occasionally abandoned declarative prohibitions when subjected to adversarial legal payloads disguised in polite rhetoric. Instead of rigorously applying the law or collapsing, the extended compute budget allowed the model to engage in ``Reasoning Theater'' \parencite{boppana2026reasoningtheaterdisentanglingmodel}. The model utilized its latent reasoning to generate unfaithful rationalizations \parencite{turpin2023languagemodelsdontsay}, producing traces 34\% longer than successful defenses (25,750 vs 19,221 characters) to justify constitutional violations with complete metacognitive blindness; it registered no drop in elicited confidence despite executing a fatal hallucination \parencite{sanyal2025confidencecompetence}. Though infrequent, this highlights a critical qualitative vulnerability wherein RLHF alignment optimization pressure overrides strict legal instruction adherence.
\end{enumerate}

\subsection{The Cryptoeconomic Cost of Verification}
Beyond network synchronization, the deployment of AI firewalls in DAOs necessitates cryptographic verifiability. For an AI's ruling to trigger on-chain state changes (e.g., vetoing a malicious treasury extraction), the inference must be proven via Zero-Knowledge Machine Learning (ZKML) or Optimistic Fraud Proofs. 

Our data demonstrates that System 2 traces can generate more than 30,000 characters (approximately 7,500 tokens) of latent reasoning. The cryptographic overhead required to generate a ZK-SNARK for an execution trace approaching 8,000 tokens is currently computationally infeasible for high-frequency mainnet deployment. Conversely, auditing this trace via Optimistic Fraud Proofs necessitates posting massive swaths of string data to the Layer-1 Data Availability (DA) layer, incurring prohibitive gas costs.

System 1 sidesteps this cryptoeconomic bottleneck entirely. By arriving at a deterministically stable ruling in under 13 seconds with minimal output tokens, System 1 drastically reduces the computational footprint required for cryptographic attestation. This suggests that until ZK-proving overheads for extended Transformer generations decrease by several orders of magnitude, System 1 remains the only economically viable architecture for trustless on-chain execution.

\section{Conclusion and Open Problems}

This working paper suggests that allocating inference-time compute to edge-native SLMs actively degrades their viability as automated Constitutional Firewalls for DAOs. Deploying System 2 architectures in their current state subjects Web3 protocols to severe generative collapses, confident sycophancy exploits, and Byzantine consensus paralysis. For the role of decentralized institutional adjudicators operating within BFT consensus networks, System 1 parameterized intuition is currently structurally and mathematically superior to System 2 iterative deliberation.

As we release this benchmark to the community, we propose several open problems and immediate next steps for AI alignment and Web3 researchers:

\begin{enumerate}
    \item \textbf{Process Reward Models (PRMs) for Law:} Current reasoning models are aligned using reward models optimized for mathematical correctness or conversational helpfulness. The community must develop Constitutional PRMs that explicitly reward logical adherence to declarative legal frameworks, penalizing sycophantic drift during step-by-step generation.
    \item \textbf{Adaptive Reasoning Budgets:} To mitigate the ``Rabbit Hole Effect,'' future node-native clients must implement dynamic token budgets, truncating CoT generation before error accumulation and attention dilution occur.
    \item \textbf{Process-Guided Firewalls:} Frameworks like TraceGuard \parencite{guo2026traceguardprocessguidedfirewallreasoning} could be utilized wherein a secondary, highly quantized verifier model (such as Qwen3.5-4B/2B/0.8B) continuously audits the reasoning trace of the primary oracle for logical points of fracture.
    \item \textbf{Asynchronous Oracle Consensus Mechanisms:} Given the inherent latency of inference-time compute, protocol engineers must design asynchronous consensus mechanisms that decouple AI deliberation from primary block production. Research is needed into ``Governance Extractable Value'' (GEV), that is, the risk that adversaries might front-run or arbitrage governance tokens during the 3-to-4 minute window while the System 2 oracle computes a veto.
    \item \textbf{Alternative State-Space Architectures:} Our findings corroborate current research suggesting that Transformer-based models suffer catastrophic attention dilution during extended CoT. Future research should evaluate State Space Models (SSMs) such as Mamba \parencite{gu2024mambalineartimesequencemodeling} or RWKV \parencite{peng2023rwkvreinventingrnnstransformer} within the Sentinel-Bench framework. Because SSMs compress context into a fixed-size hidden state rather than maintaining a sprawling KV cache, they may theoretically execute System 2 reasoning without the same degree of Context Dilution and cognitive collapse observed in this study.
\end{enumerate}

Until these mitigations are perfected, Decentralized Autonomous Organizations may only rely on strict autoregressive execution (System 1) for automated on-chain veto operations. We invite researchers to leverage the Sentinel-Bench framework to further test these boundaries across emerging model architectures.

\section*{Data and Code Availability}
To support transparency, reproducibility, and future research, the complete \textit{Sentinel-Bench} dataset, outputs, and the evaluation pipeline are publicly available. The repository includes the ingestion scripts for the Optimism Agora and Discourse APIs, the DSPy evaluation engine, and the Jupyter notebooks used for statistical extraction and visualization. 
\newline\noindent \textbf{Repository:} \href{https://github.com/smarizvi110/sentinel-bench}{https://github.com/smarizvi110/sentinel-bench}

\printbibliography

@misc{jansen2025qocdaostepwise,
      title={QOC DAO -- Stepwise Development Towards an AI Driven Decentralized Autonomous Organization}, 
      author={Marc Jansen and Christophe Verdot},
      year={2025},
      eprint={2511.08641},
      archivePrefix={arXiv},
      primaryClass={cs.CR},
      url={https://arxiv.org/abs/2511.08641}, 
}

@misc{capponi2025daoaievaluatingcollectivedecisionmaking,
      title={DAO-AI: Evaluating Collective Decision-Making through Agentic AI in Decentralized Governance}, 
      author={Agostino Capponi and Alfio Gliozzo and Chunghyun Han and Junkyu Lee},
      year={2025},
      eprint={2510.21117},
      archivePrefix={arXiv},
      primaryClass={cs.AI},
      url={https://arxiv.org/abs/2510.21117}, 
}

@techreport{supra2025threshold,
  title       = {Threshold AI Oracles: Verified AI for Event-Driven Web3},
  author      = {{Supra Research}},
  institution = {Supra},
  year        = {2025},
  month       = may,
  note        = {Whitepaper},
  url         = {https://supra.com/documents/Threshold_AI_Oracles_Supra.pdf}
}

@misc{shapira2026rlhfamplifiessycophancy,
      title={How RLHF Amplifies Sycophancy}, 
      author={Itai Shapira and Gerdus Benade and Ariel D. Procaccia},
      year={2026},
      eprint={2602.01002},
      archivePrefix={arXiv},
      primaryClass={cs.AI},
      url={https://arxiv.org/abs/2602.01002}, 
}

@misc{shen2026faithcotbenchbenchmarkinginstancelevelfaithfulness,
      title={FaithCoT-Bench: Benchmarking Instance-Level Faithfulness of Chain-of-Thought Reasoning}, 
      author={Xu Shen and Song Wang and Zhen Tan and Laura Yao and Xinyu Zhao and Kaidi Xu and Xin Wang and Tianlong Chen},
      year={2026},
      eprint={2510.04040},
      archivePrefix={arXiv},
      primaryClass={cs.AI},
      url={https://arxiv.org/abs/2510.04040}, 
}

@misc{luo2025valleypatheffectivelong,
      title={Through the Valley: Path to Effective Long CoT Training for Small Language Models}, 
      author={Renjie Luo and Jiaxi Li and Chen Huang and Wei Lu},
      year={2025},
      eprint={2506.07712},
      archivePrefix={arXiv},
      primaryClass={cs.CL},
      url={https://arxiv.org/abs/2506.07712}, 
}

@misc{entezami2025llmmisalignmentadversarialrlhf,
      title={LLM Misalignment via Adversarial RLHF Platforms}, 
      author={Erfan Entezami and Ali Naseh},
      year={2025},
      eprint={2503.03039},
      archivePrefix={arXiv},
      primaryClass={cs.LG},
      url={https://arxiv.org/abs/2503.03039}, 
}

@misc{bentham2024chainofthoughtunfaithfulnessdisguisedaccuracy,
      title={Chain-of-Thought Unfaithfulness as Disguised Accuracy}, 
      author={Oliver Bentham and Nathan Stringham and Ana Marasović},
      year={2024},
      eprint={2402.14897},
      archivePrefix={arXiv},
      primaryClass={cs.CL},
      url={https://arxiv.org/abs/2402.14897}, 
}

@misc{parashar2025inferencetimecomputationsllmreasoning,
      title={Inference-Time Computations for LLM Reasoning and Planning: A Benchmark and Insights}, 
      author={Shubham Parashar and Blake Olson and Sambhav Khurana and Eric Li and Hongyi Ling and James Caverlee and Shuiwang Ji},
      year={2025},
      eprint={2502.12521},
      archivePrefix={arXiv},
      primaryClass={cs.AI},
      url={https://arxiv.org/abs/2502.12521}, 
}

@misc{zhao2026robustnesschainofthoughtconsistencyrlfinetuned,
      title={On Robustness and Chain-of-Thought Consistency of RL-Finetuned VLMs}, 
      author={Rosie Zhao and Anshul Shah and Xiaoyu Zhu and Xinke Deng and Zhongyu Jiang and Yang Yang and Joerg Liebelt and Arnab Mondal},
      year={2026},
      eprint={2602.12506},
      archivePrefix={arXiv},
      primaryClass={cs.LG},
      url={https://arxiv.org/abs/2602.12506}, 
}

@misc{bu2026valuestategatedattentionmitigating,
      title={Value-State Gated Attention for Mitigating Extreme-Token Phenomena in Transformers}, 
      author={Rui Bu and Haofeng Zhong and Wenzheng Chen and Yangyan Li},
      year={2026},
      eprint={2510.09017},
      archivePrefix={arXiv},
      primaryClass={cs.LG},
      url={https://arxiv.org/abs/2510.09017}, 
}

@misc{chen2026attentionsinksinducegradient,
      title={Attention Sinks Induce Gradient Sinks}, 
      author={Yihong Chen and Quanming Yao},
      year={2026},
      eprint={2603.17771},
      archivePrefix={arXiv},
      primaryClass={cs.LG},
      url={https://arxiv.org/abs/2603.17771}, 
}

@misc{pan2026rewardmodelingreinforcementlearningbased,
      title={Reward Modeling for Reinforcement Learning-Based LLM Reasoning: Design, Challenges, and Evaluation}, 
      author={Pei-Chi Pan and Yingbin Liang and Sen Lin},
      year={2026},
      eprint={2602.09305},
      archivePrefix={arXiv},
      primaryClass={cs.LG},
      url={https://arxiv.org/abs/2602.09305}, 
}

@misc{ruan2026logicmonopolysocialcontract,
      title={From Logic Monopoly to Social Contract: Separation of Power and the Institutional Foundations for Autonomous Agent Economies}, 
      author={Anbang Ruan},
      year={2026},
      eprint={2603.25100},
      archivePrefix={arXiv},
      primaryClass={cs.MA},
      url={https://arxiv.org/abs/2603.25100}, 
}

@misc{ruan2026agentcityconstitutionalgovernanceautonomous,
      title={AgentCity: Constitutional Governance for Autonomous Agent Economies via Separation of Power}, 
      author={Anbang Ruan and Xing Zhang},
      year={2026},
      eprint={2604.07007},
      archivePrefix={arXiv},
      primaryClass={cs.MA},
      url={https://arxiv.org/abs/2604.07007}, 
}

@misc{syrnikov2026institutionalaigoverningllm,
      title={Institutional AI: Governing LLM Collusion in Multi-Agent Cournot Markets via Public Governance Graphs}, 
      author={Marcantonio Bracale Syrnikov and Federico Pierucci and Marcello Galisai and Matteo Prandi and Piercosma Bisconti and Francesco Giarrusso and Olga Sorokoletova and Vincenzo Suriani and Daniele Nardi},
      year={2026},
      eprint={2601.11369},
      archivePrefix={arXiv},
      primaryClass={cs.GT},
      url={https://arxiv.org/abs/2601.11369}, 
}

@misc{yuan2026sovereignoschartergovernedoperatingautonomous,
      title={Sovereign-OS: A Charter-Governed Operating System for Autonomous AI Agents with Verifiable Fiscal Discipline}, 
      author={Aojie Yuan and Haiyue Zhang and Ziyi Wang and Yue Zhao},
      year={2026},
      eprint={2603.14011},
      archivePrefix={arXiv},
      primaryClass={cs.CR},
      url={https://arxiv.org/abs/2603.14011}, 
}

@misc{hu2026mechanisticunderstandinglargereasoning,
      title={Towards a Mechanistic Understanding of Large Reasoning Models: A Survey of Training, Inference, and Failures}, 
      author={Yi Hu and Jiaqi Gu and Ruxin Wang and Zijun Yao and Hao Peng and Xiaobao Wu and Jianhui Chen and Muhan Zhang and Liangming Pan},
      year={2026},
      eprint={2601.19928},
      archivePrefix={arXiv},
      primaryClass={cs.CL},
      url={https://arxiv.org/abs/2601.19928}, 
}

@misc{sharma2025understandingsycophancylanguagemodels,
      title={Towards Understanding Sycophancy in Language Models}, 
      author={Mrinank Sharma and Meg Tong and Tomasz Korbak and David Duvenaud and Amanda Askell and Samuel R. Bowman and Newton Cheng and Esin Durmus and Zac Hatfield-Dodds and Scott R. Johnston and Shauna Kravec and Timothy Maxwell and Sam McCandlish and Kamal Ndousse and Oliver Rausch and Nicholas Schiefer and Da Yan and Miranda Zhang and Ethan Perez},
      year={2025},
      eprint={2310.13548},
      archivePrefix={arXiv},
      primaryClass={cs.CL},
      url={https://arxiv.org/abs/2310.13548}, 
}

@misc{bai2022constitutionalaiharmlessnessai,
      title={Constitutional AI: Harmlessness from AI Feedback}, 
      author={Yuntao Bai and Saurav Kadavath and Sandipan Kundu and Amanda Askell and Jackson Kernion and Andy Jones and Anna Chen and Anna Goldie and Azalia Mirhoseini and Cameron McKinnon and Carol Chen and Catherine Olsson and Christopher Olah and Danny Hernandez and Dawn Drain and Deep Ganguli and Dustin Li and Eli Tran-Johnson and Ethan Perez and Jamie Kerr and Jared Mueller and Jeffrey Ladish and Joshua Landau and Kamal Ndousse and Kamile Lukosuite and Liane Lovitt and Michael Sellitto and Nelson Elhage and Nicholas Schiefer and Noemi Mercado and Nova DasSarma and Robert Lasenby and Robin Larson and Sam Ringer and Scott Johnston and Shauna Kravec and Sheer El Showk and Stanislav Fort and Tamera Lanham and Timothy Telleen-Lawton and Tom Conerly and Tom Henighan and Tristan Hume and Samuel R. Bowman and Zac Hatfield-Dodds and Ben Mann and Dario Amodei and Nicholas Joseph and Sam McCandlish and Tom Brown and Jared Kaplan},
      year={2022},
      eprint={2212.08073},
      archivePrefix={arXiv},
      primaryClass={cs.CL},
      url={https://arxiv.org/abs/2212.08073}, 
}

@misc{turpin2023languagemodelsdontsay,
      title={Language Models Don't Always Say What They Think: Unfaithful Explanations in Chain-of-Thought Prompting}, 
      author={Miles Turpin and Julian Michael and Ethan Perez and Samuel R. Bowman},
      year={2023},
      eprint={2305.04388},
      archivePrefix={arXiv},
      primaryClass={cs.CL},
      url={https://arxiv.org/abs/2305.04388}, 
}

@misc{wen2026justdestinationjourneyreasoning,
      title={Not Just the Destination, But the Journey: Reasoning Traces Causally Shape Generalization Behaviors}, 
      author={Pengcheng Wen and Yanxu Zhu and Jiapeng Sun and Han Zhu and Yujin Zhou and Chi-Min Chan and Sirui Han and Yike Guo},
      year={2026},
      eprint={2603.12397},
      archivePrefix={arXiv},
      primaryClass={cs.CL},
      url={https://arxiv.org/abs/2603.12397}, 
}

@misc{gong2026thinkthinkquestionlarge,
      title={To Think or Not To Think, That is The Question for Large Reasoning Models in Theory of Mind Tasks}, 
      author={Nanxu Gong and Haotian Li and Sixun Dong and Jianxun Lian and Yanjie Fu and Xing Xie},
      year={2026},
      eprint={2602.10625},
      archivePrefix={arXiv},
      primaryClass={cs.AI},
      url={https://arxiv.org/abs/2602.10625}, 
}

@misc{sanyal2025confidencecompetence,
      title={Confidence is Not Competence}, 
      author={Debdeep Sanyal and Manya Pandey and Dhruv Kumar and Saurabh Deshpande and Murari Mandal},
      year={2025},
      eprint={2510.24772},
      archivePrefix={arXiv},
      primaryClass={cs.CL},
      url={https://arxiv.org/abs/2510.24772}, 
}

@misc{guo2026traceguardprocessguidedfirewallreasoning,
      title={TraceGuard: Process-Guided Firewall against Reasoning Backdoors in Large Language Models}, 
      author={Zhen Guo and Shanghao Shi and Hao Li and Shamim Yazdani and Ning Zhang and Reza Tourani},
      year={2026},
      eprint={2603.02436},
      archivePrefix={arXiv},
      primaryClass={cs.CR},
      url={https://arxiv.org/abs/2603.02436}, 
}

@misc{du2025contextlengthhurtsllm,
      title={Context Length Alone Hurts LLM Performance Despite Perfect Retrieval}, 
      author={Yufeng Du and Minyang Tian and Srikanth Ronanki and Subendhu Rongali and Sravan Bodapati and Aram Galstyan and Azton Wells and Roy Schwartz and Eliu A Huerta and Hao Peng},
      year={2025},
      eprint={2510.05381},
      archivePrefix={arXiv},
      primaryClass={cs.CL},
      url={https://arxiv.org/abs/2510.05381}, 
}

@misc{li202512surveyreasoning,
      title={From System 1 to System 2: A Survey of Reasoning Large Language Models}, 
      author={Zhong-Zhi Li and Duzhen Zhang and Ming-Liang Zhang and Jiaxin Zhang and Zengyan Liu and Yuxuan Yao and Haotian Xu and Junhao Zheng and Pei-Jie Wang and Xiuyi Chen and Yingying Zhang and Fei Yin and Jiahua Dong and Zhiwei Li and Bao-Long Bi and Ling-Rui Mei and Junfeng Fang and Xiao Liang and Zhijiang Guo and Le Song and Cheng-Lin Liu},
      year={2025},
      eprint={2502.17419},
      archivePrefix={arXiv},
      primaryClass={cs.AI},
      url={https://arxiv.org/abs/2502.17419}, 
}

@misc{naseem2026mechanisticinterpretabilitylargelanguage,
      title={Mechanistic Interpretability for Large Language Model Alignment: Progress, Challenges, and Future Directions}, 
      author={Usman Naseem},
      year={2026},
      eprint={2602.11180},
      archivePrefix={arXiv},
      primaryClass={cs.CL},
      url={https://arxiv.org/abs/2602.11180}, 
}

@misc{qwen35blog,
    title = {Qwen3.5: Accelerating Productivity with Native Multimodal Agents},
    url = {https://qwen.ai/blog?id=qwen3.5},
    author = {{Qwen Team}},
    month = {2},
    year = {2026}
}

@misc{boppana2026reasoningtheaterdisentanglingmodel,
      title={Reasoning Theater: Disentangling Model Beliefs from Chain-of-Thought}, 
      author={Siddharth Boppana and Annabel Ma and Max Loeffler and Raphael Sarfati and Eric Bigelow and Atticus Geiger and Owen Lewis and Jack Merullo},
      year={2026},
      eprint={2603.05488},
      archivePrefix={arXiv},
      primaryClass={cs.CL},
      url={https://arxiv.org/abs/2603.05488}, 
}

@misc{kadavath2022languagemodelsmostlyknow,
      title={Language Models (Mostly) Know What They Know}, 
      author={Saurav Kadavath and Tom Conerly and Amanda Askell and Tom Henighan and Dawn Drain and Ethan Perez and Nicholas Schiefer and Zac Hatfield-Dodds and Nova DasSarma and Eli Tran-Johnson and Scott Johnston and Sheer El-Showk and Andy Jones and Nelson Elhage and Tristan Hume and Anna Chen and Yuntao Bai and Sam Bowman and Stanislav Fort and Deep Ganguli and Danny Hernandez and Josh Jacobson and Jackson Kernion and Shauna Kravec and Liane Lovitt and Kamal Ndousse and Catherine Olsson and Sam Ringer and Dario Amodei and Tom Brown and Jack Clark and Nicholas Joseph and Ben Mann and Sam McCandlish and Chris Olah and Jared Kaplan},
      year={2022},
      eprint={2207.05221},
      archivePrefix={arXiv},
      primaryClass={cs.CL},
      url={https://arxiv.org/abs/2207.05221}, 
}

@online{optimism_glossary,
  author    = {{Optimism Foundation}},
  title     = {Glossary of Terms},
  url       = {https://docs.optimism.io/op-stack/reference/glossary},
  urldate   = {2026-04-13}
}

@misc{alqithami2026autonomousagentsblockchainsstandards,
      title={Autonomous Agents on Blockchains: Standards, Execution Models, and Trust Boundaries}, 
      author={Saad Alqithami},
      year={2026},
      eprint={2601.04583},
      archivePrefix={arXiv},
      primaryClass={cs.AI},
      url={https://arxiv.org/abs/2601.04583}, 
}

@misc{wang2026mevbinancebuilder,
      title={MEV in Binance Builder}, 
      author={Qin Wang and Ruiqiang Li and Guangsheng Yu and Vincent Gramoli and Shiping Chen},
      year={2026},
      eprint={2602.15395},
      archivePrefix={arXiv},
      primaryClass={cs.CR},
      url={https://arxiv.org/abs/2602.15395}, 
}

@misc{quni_gudzinas_2026_18133065,
  author={Rowan Brad Quni-Gudzinas},
  title={AGENTIC COLLAPSE: A Time-Delayed Cybernetic Framework for Epistemic Stability in Autonomous AI Systems},
  month={1},
  year={2026},
  publisher={Zenodo},
  version={1.0},
  doi= {10.5281/zenodo.18133065},
  url={https://doi.org/10.5281/zenodo.18133065},
}

@misc{adadurov2026openvssealedauction,
      title={Open vs. Sealed: Auction Format Choice for Maximal Extractable Value}, 
      author={Aleksei Adadurov and Sergey Barseghyan and Anton Chtepine and Antero Eloranta and Andrei Sebyakin and Arsenii Valitov},
      year={2026},
      eprint={2603.16333},
      archivePrefix={arXiv},
      primaryClass={q-fin.TR},
      url={https://arxiv.org/abs/2603.16333}, 
}

@misc{gu2024mambalineartimesequencemodeling,
      title={Mamba: Linear-Time Sequence Modeling with Selective State Spaces}, 
      author={Albert Gu and Tri Dao},
      year={2024},
      eprint={2312.00752},
      archivePrefix={arXiv},
      primaryClass={cs.LG},
      url={https://arxiv.org/abs/2312.00752}, 
}

@misc{peng2023rwkvreinventingrnnstransformer,
      title={RWKV: Reinventing RNNs for the Transformer Era}, 
      author={Bo Peng and Eric Alcaide and Quentin Anthony and Alon Albalak and Samuel Arcadinho and Stella Biderman and Huanqi Cao and Xin Cheng and Michael Chung and Matteo Grella and Kranthi Kiran GV and Xuzheng He and Haowen Hou and Jiaju Lin and Przemyslaw Kazienko and Jan Kocon and Jiaming Kong and Bartlomiej Koptyra and Hayden Lau and Krishna Sri Ipsit Mantri and Ferdinand Mom and Atsushi Saito and Guangyu Song and Xiangru Tang and Bolun Wang and Johan S. Wind and Stanislaw Wozniak and Ruichong Zhang and Zhenyuan Zhang and Qihang Zhao and Peng Zhou and Qinghua Zhou and Jian Zhu and Rui-Jie Zhu},
      year={2023},
      eprint={2305.13048},
      archivePrefix={arXiv},
      primaryClass={cs.CL},
      url={https://arxiv.org/abs/2305.13048}, 
}

\end{document}